# Hybrid gene selection approach using XGBoost and multi-objective genetic algorithm for cancer classification

Xiongshi Deng[1,2] · Min Li[1,2,*] · Shaobo Deng [1,2] · Lei Wang [1,2]

[1] *School of Information Engineering, Nanchang Institute of Technology, Jiangxi 330099, P.R. China*

[2] *Jiangxi Province Key Laboratory of Water Information Cooperative Sensing and Intelligent Processing, Jiangxi 330099, P.R. China*

**Abstract**

Microarray gene expression data are often accompanied by a large number of genes and a small number of samples. However, only a few of these genes are relevant to cancer, resulting in significant gene selection challenges. Hence, we propose a two-stage gene selection approach by combining extreme gradient boosting (XGBoost) and a multi-objective optimization genetic algorithm (XGBoost-MOGA) for cancer classification in microarray datasets. In the first stage, the genes are ranked use an ensemble-based feature selection using XGBoost. This stage can effectively remove irrelevant genes and yield a group comprising the most relevant genes related to the class. In the second stage, XGBoost-MOGA searches for an optimal gene subset based on the most relevant genes' group using a multi-objective optimization genetic algorithm. We performed comprehensive experiments to compare XGBoost-MOGA with other state-of-the-art feature selection methods using two well-known learning classifiers on 13 publicly available microarray expression datasets. The experimental results show that XGBoost–MOGA yields significantly better results than previous state-of-the-art algorithms in terms of various evaluation criteria, such as accuracy, F-score, precision, and recall.

**Keywords** XGBoost · Feature selection · Microarray gene expression · Multi-objective Genetic Algorithm · Classification

## 1 Introduction

Since the introduction of gene chip technology, a significant amount of microarray gene expression information data, which conceal valuable biological information, have been generated [1, 2]. The analysis of these data and the mining of potential biological information provide a new condition for diagnosing and treating complex diseases [3]. However, genetic datasets generally involve a small number of samples, high dimensions, imbalanced distribution, and considerable noise [4–8]. Publicly available microarray datasets for gene expression generally contain more than 2,000 genes, whereas some contain more than 20,000 genes [3, 9, 10]. Most of these genes are unrelated to disease. Directly training these large numbers of genetic features in the model is time consuming and does not avoid the curse of dimensionality [11]. Consequently, diagnosis and treatment based on genetic data will be extremely challenging. It is challenging to effectively identify a few genes among thousands of genes that are associated with a disease [12].

Feature (gene) selection is also known as attribute selection. It selects a feature with strong correlation from the original dataset to reduce the dimensions of the dataset [13, 14]. Feature selection is critical in machine learning and data mining preprocessing. For the design of feature selection, four types of standard feature selection methods exist: filter, wrapper, embedded, and hybrid(ensemble) methods [2, 12, 15–18].

Filter methods are a preprocessing process to filter out irrelevant features independent of the subsequent learning process. It is based on the analysis of the sample data's intrinsic properties used to train the classifier. Statistical information, such as the correlation level between features and classes, correlation level between different features, distance between training samples, and correlation consistency, can be used as a basis to evaluate the merits and demerits of feature subsets [18, 19]. Each feature's importance is calculated separately based on the class differentiation level of each feature, and a ranking table depicting the importance of the feature is obtained. Subsequently, the top N features are selected to form the targeted subset of selected features. Currently, the most typically used filter methods include chi-square statistics, information gain, Relief, and correlation measurement [20–23].

Wrapper methods evaluate the feature subset using an induction algorithm as a black box based on the predictive power of the induction algorithm [15, 17, 18]. When selecting the optimal feature subset, the training dataset must be further segregated into a subtraining set and a subtest set. The selected feature subset is directly used to train the classifier on the subtraining set and during training. The prediction accuracy of the model on the subtest set is used as a basis for evaluating the advantages and disadvantages of the feature subset. To construct a classification model using a wrapper method, the feature subset with the best classification performance is selected to obtain a higher classification performance model. As such, wrapper methods offer excellent classification performances. To some extent, embedded methods are similar to wrapper methods. The embedded method optimizes an objective function to achieve feature selection during the classifier's training [4]. In this process, different feature combinations can be formed. Compared with the wrapper method, the embedded method does not require the repeated segregation of the training dataset into subtraining and subtest sets, thereby avoiding a loop and the repeated training of the learning machine for evaluating each feature subset. Therefore, it can quickly select the optimal feature subset.

Hybrid methods are generally formed based on the joint mixing of filter method, wrapper method and embedded method, as well as the improvement and expansion on the basis of these three methods, which good integrate the advantage of above three methods. For instance, in [23], Ghosh et al. proposed a 2-stage model for feature selection in microarray datasets. This approach first develops an ensemble of filter methods by considering the union and intersection of the top-n features of three filter methods, and then uses genetic algorithm (GA) on the union and intersection to get the fine-tuned results. Experimental results show that the model has superiority compared with state-of-the-art methods on 5 publicly available cancer data sets. Shukla et al. proposed a novel hybrid feature selection algorithm, called Filter-Wrapper Feature Selection (FWFS), to select cancer-related genes from high-dimensional microarray data [19]. This method first uses Conditional Mutual Information Maximization (CMIM) [24] to select the top-ranked genes, and then uses the Binary Genetic Algorithm (BGA) to select the optimal subset of genes. In [25], Lu et al. proposed a hybrid feature selection algorithm that combines the mutual information maximization (MIM) and the adaptive genetic algorithm (AGA), to reduce the dimensionality of gene expression data and improve classification accuracy.

Most previous relevant studies primarily focused on filter and wrapper methods for the three feature selection methods mentioned above. Many hybrid gene selection methods based on the filter-wrapper have been proposed to improve the classification accuracy of models; however, research regarding embedded strategies is scarce. Because eXtreme Gradient Boosting (XGBoost) [26] is an ensemble learning algorithm improved based on the Gradient Boosting Decision Tree (GBDT) [27], it does not necessitate feature engineering, which is highly advantageous. However, most studies regarding XGBoost focuses on adjusting its parameters for model training, disregarding the fact that XGBoost can also perform feature selection. Therefore, to select the most relevant genes from a microarray dataset, improve the model performance, reduce the computing time, and increase the diversity of research, we herein propose a hybrid gene selection approach using XGBoost and a multi-objective GA, named

XGBoost-MOGA. This approach first ranks gene importance based on XGBoost, eliminates low-ranking genes, and obtains the initial gene subset. Subsequently, the obtained gene subset is input to the MOGA to obtain the optimal gene subset. Based on these two gene selection steps, the optimal gene subset can be selected from the original dataset. A comparison and an analysis of experimental results verified the advantage of XGBoost-MOGA in terms of the number of selected genes and execution time, benefitting particularly classification performance.

The remainder of this paper is organized as follows. In Section 2, we review the research work related to XGBoost and genetic algorithms. Section 3 briefly reviews the principle of XGBoost and MOGA and then presents XGBoost-MOGA in detail. In Section 4, we compare the XGBoost-MOGA approach with two feature selection algorithms that are the most related to it and other state-of-the-art feature selection algorithms on 13 publicly available microarray expression datasets. Section 5 presents the conclusions and future work.

## 2 Related work

In this section, a brief overview of some recent works on feature selection and classification in microarray datasets, especially in terms of XGBoost and genetic algorithms.

In [28], in order to more accurate to improve primary lesions' diagnostic efficiency, Chen et al. first use two feature selection methods to select genes with high expression levels from the data set, and then use XGBoost for classification. In [29], in order to achieve higher accuracy in the public colon cancer gene expression data set, Islam et al. proposed an ensemble method, which combines multiple classification methods. This method first uses Linear Discriminant Analysis (LDA) and Principle Component Analysis (PCA) to cope up with the high dimensionality of the data, and then uses an ensemble learning model with k-Nearest Neighbors (kNN), Random Forest (RF), Kernel Support Vector Machines (KSVM), XGBoost, and Bayes Generalized Linear Model (GLM) to classification.

In [9], Li et al. designed an algorithm for predicting gene expression values based on XGBoost, which integrates multiple tree models and has stronger interpretability. This XGBoost model is compared with other existing models on the GEO dataset and RNA-seq dataset and has achieved significant advantages in predicting gene expression values. In [30],

Dimitrakopoulos et al. proposed a classification scheme based on XGBoost, a recent tree-based classification algorithm, to detect the most discriminative pathways related with a disease. This method is verified on an aging gene expression data set, proving that XGBoost is superior to other well-known classification methods in biological data.

To identify key MiRNAs and clinical features for improving the accuracy of ovarian cancer recurrence prediction, Sujamol et al. proposed a multistage feature selection methodology, called Inheritable Bi-objective Combinatorial Genetic Algorithm (IBCGA) [31]. From 588 MiRNAs, 6 key MiRNAs were selected using the IBCGA followed by factor analysis. Seven classifiers are used to classify and predict the selected MiRNA genes, and among which the XGBoost classifier has the best performance, with a prediction accuracy of 91.86%.

To enhance lung cancer recurrence prediction and reduce painful treatment, in [32], Abdu-Aljabar et al. used the XGBoost model to enhance prediction and detection, and used gene expression and cancer transcript changes to detect lung cancer diagnosis and predict its recurrence after surgery.

**3 Hybrid Gene Selection Approach**

In this section, a hybrid gene selection approach named XGBoost-MOGA is introduced. The XGBoost-MOGA approach combines an embedded method, i.e., XGBoost, and a wrapper method, i.e., the MOGA. In Sections 3.1 and 3.2, we present the principles of XGBoost and MOGA, respectively. In Section 3.3, we present the procedure of XGBoost-MOGA in detail.

3.1 XGBoost (eXtreme Gradient Boosting)

XGBoost, which was introduced in February 2014, is a machine learning library that focuses on gradient boosting algorithms. The author of XGBoost is Chen Tianqi, an expert in machine learning at the University of Washington [26]. XGBoost is essentially the integration of multiple CARTs (Classification and Regression Trees). The CART is a type of decision tree that can be used for both classification and regression tasks. Therefore, XGBoost is an ensemble learning algorithm that is improved based on the GBDT; it does not require feature engineering, which is highly advantageous. The traditional GBDT objective function superimposes the residual tree of different rounds after iteration to predict the target classes. Simultaneously, XGBoost improves the traditional GBDT objective function by adding a regular term based on the original function to reduce the possibility of overfitting and accelerate convergence [9, 26, 33]. XGBoost can automatically obtain the importance of features such that features can be filtered. Therefore, we used XGBoost to select the most relevant features during the initial feature selection.

In Eq. (1), $L(y_i, \hat{y}_i)$ is the loss of the square deviation as a function of the actual value $y_i$ and predictive value $\hat{y}_i$, and $\Omega(f_i)$ is a regularization term. In Eq. (2), $\gamma$ is the difficulty coefficient of the tree split for controlling the tree's generation. $T$ is the number of leaf nodes, and $\lambda$ is the $l2$ regularity coefficient.

$$Obj(t) = \sum_{i=1}^{n} L(y_i, \hat{y}_i) + \sum_{i=1}^{t} \Omega(f_i) \tag{1}$$

$$\Omega(f_t) = \gamma T + \frac{1}{2}\lambda \sum_{i=1}^{T} \omega_j^2 \tag{2}$$

After $t$ iterations, the predicted value can be written as

$$\hat{y}_i^{(t)} = \hat{y}_i^{(t-1)} + f_t(x_i) \tag{3}$$

Subsequently, the objective function can be updated to

$$Obj(t) = \sum_{i=1}^{n} L(y_i, \hat{y}^{(t-1)} + f_t(x_i)) + \Omega(f_t) \tag{4}$$

XGBoost differs from the GBDT. It can expand the loss function based on the second derivative of Taylor's formula such that the new objective function obtained will have a faster convergence rate and higher accuracy than the original GBDT. The final objective function is expressed as shown in Eq. (5).

$$Obj^{(t)} \cong \sum_{i=1}^{n} [L(y_i, \hat{y}^{(t-1)}) + g_i f_t(x_i) + \frac{1}{2} h_i f_t^2(x_i)] + \Omega(f_t) \tag{5}$$

$$g_i = \frac{\partial(L(y_i, \hat{y}^{(t-1)}))}{\partial \hat{y}^{(t-1)}} \tag{6}$$

$$h_i = \frac{\partial^2(L(y_i, \hat{y}^{(t-1)}))}{\partial \hat{y}^{(t-1)}}, \tag{7}$$

where $g_i$ and $h_i$ are the first and second derivatives of the loss function $L(\phi)$, respectively.

3.2 Multi-objective genetic algorithm (MOGA)

The GA [25] is a natural adaptive optimization method that simulates the problem to be solved as a biological evolution process [19]. It generates next-generation solutions through duplication, crossover, and mutation and gradually eliminates solutions with low fitness function values [34, 35]. The GA can use the search information of multiple search points simultaneously. The use of probabilistic search technology can prevent iterative information points falling into the local optimal solution. The GA is robust, which search flexibility is excellent, and it can yield the global optimal solution rapidly [19, 35]. It has been widely used in various complex optimization problems. Multi-objective optimization problem is a kind of complex and difficult to solve optimization problem. Therefore, researchers have developed a variety of solution methods, among which using genetic algorithm to solve multi-objective optimization problems is a very effective method.

The multi-objective optimization problem is to solve the optimization problem of multiple objective functions under certain constrained conditions, which can be defined formally as follows:

Let $O = (X, Y, F)$ be an initial selected dataset, where $X$ is the sample set, $Y$ is the class set, and $F = \{f_1, f_2, ..., f_m\}$ is the feature set.

In multi-objective optimization problems, there are usually conflicts between multiple objectives, and the solution is not unique, but exists in an optimal solution set. Hence, we need to find out a set of features $S = \{f_1, f_2, ..., f_k\} \in F (k \leq m)$ from the optimal solution set. In addition, the number of features in the selected feature subset should be as small as possible.

In this study, the MOGA is a wrapper-based method that can be used to obtain the optimal feature subset. After obtaining the initial feature subset by XGBoost, we used the MOGA to obtain the optimal feature subset. Fig. 1 shows an illustration of the MOGA for feature selection.

1) Individual code

From a biological perspective, the genotype is the internal expression of trait chromosomes. The phenotype is the external expression of traits determined by chromosomes or individuals' formation based on the genotype. Therefore, the genotype determines the phenotype. In the MOGA, all the operating objects are genes (i.e., 0 and 1); therefore, the input object must be encoded into corresponding gene sequences. The input object is typically categorized into two values: discrete and constant floating-point values. First, discrete values are coded in binary, and continuous values are coded by dividing continuous cells. Subsequently, the input object is encoded based on the input object's value type. Finally, the corresponding coding gene sequence of the input object is obtained. Because the microarray dataset values are discrete values, binary coding is used to yield 0/1 as the individual's gene sequence distribution [33, 36]. In our experiment, the number of individual genes was constructed based on the number of genes obtained after initial gene selection via XGBoost.

2) Fitness value calculation

The MOGA evaluates each individual's advantages and disadvantages by the individual fitness value, thereby determining its genetic opportunity. Individuals with high fitness values have a greater probability of surviving in the next generation. The calculation of fitness values is vital for the MOGA. When using a MOGA for feature selection, the selection of the fitness function can be categorized into four types: distance-based separability, average-variance-ratio based [3], Fisher based, and classification model [36, 37]. If the classification model is adopted, then the feature-selection methods belong to the wrapper category. Considering that wrapper methods can yield excellent classification performances, the classification model is used in the fitness function. We used the nearest neighbor classification model to calculate each chromosome's fitness value, as follows [3, 36]:

$$fitness = KNN_{acc}(C_i), \tag{8}$$

where $C$ represents chromosomes, $i$ is the chromosome number, and $acc$ represents accuracy.

3) Selection operation

The purpose of selection is to select good individuals from the exchanged group such that they can reproduce and become parents to the next generation. The MOGA reflects this idea through a selection process. The principle of selection is that individuals with strong adaptability have a high probability of contributing to the next generation, and it implements Darwin's principle of survival of the fittest [34]. In our study, the tournament selection method was used to select individuals randomly [19]. The tournament method selection strategy randomly selects $N$ individuals from the population each time and then selects the best one among them to enter the progeny population. This operation is repeated until the new population size reaches the original population size.

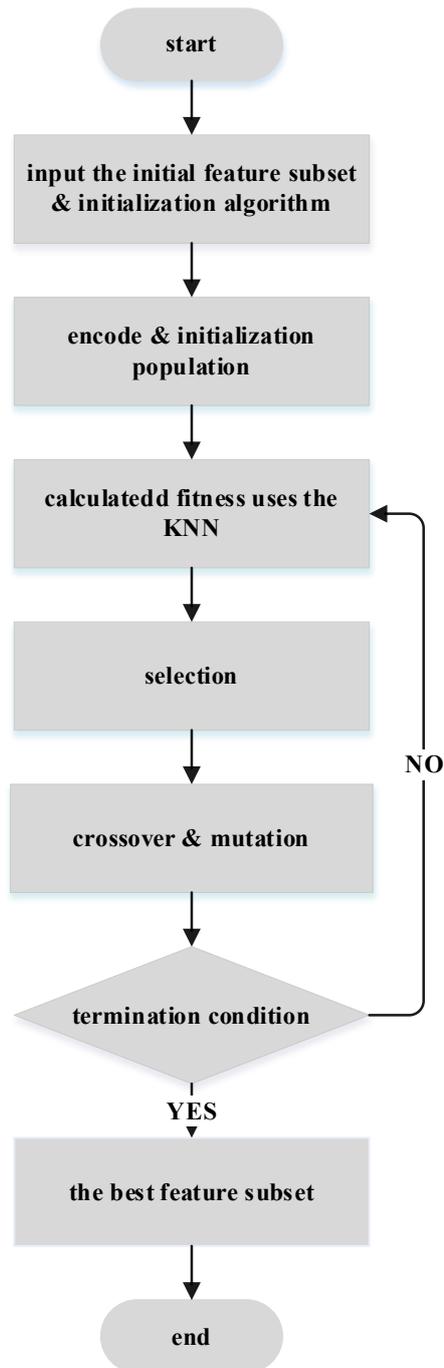

Fig.1. Flow-chart of multi-objective genetic algorithm

4) Crossover

Crossover is the primary operation of generating new individuals in a MOGA. It exchanges some chromosomes between two individuals with a certain probability. In this study, we adopted the uniform crossover method. The genes on each locus of two paired individuals are exchanged with the same crossover probability, forming two new individuals.

5) Mutation

First, a certain number of individuals are randomly selected from the population. For the individuals selected, the value of a gene in the string structure data is randomly changed with a certain probability such that 0 becomes 1 and 1 becomes 0. As in the biological world, mutation provides an opportunity for the generation of new individuals; however, the mutation rate of the MOGA is extremely low. We randomly selected a bit from the gene to perform flipping.

6) Termination condition

Generally, the MOGA comprises three termination conditions: (1) For a maximum genetic iteration MAXGEN (determined in advance), the algorithm stops when iterating to the MAXGEN times. (2) For the lower bound calculation method, when the evolution reaches the required deviation, the algorithm terminates. (3) When the monitored algorithm can no longer improve the solution's performance, the calculation is stopped. The population parameters set in the experiment were relatively large, and the population variation was considerable. Criterion (1) was adopted to obtain the most adaptable population to yield the optimal parameter performance.

3.3 XGBoost-MOGA

In this section, we propose a two-stage gene selection approach that combines XGBoost and multi-objective GA, named XGBoost-MOGA, for cancer classification in microarray datasets. In the first stage, the genes are ranked via XGBoost, and only those with scores greater than zero were retained. This stage can effectively remove irrelevant genes and yield a group of genes that are the most relevant to the class. In the second stage, XGBoost-MOGA searches for an optimal gene subset based on the most relevant genes' group using a multi-objective GA. This stage can remove redundant genes and further enhance cancer classification. Fig. 2 shows an illustration of XGBoost-MOGA for gene selection. Algorithm 1 shows an overview of the overall framework of the proposed approach. Tables 1–2 are the parameters of XGBoost and MOGA, respectively.

| Algorithm 1. Proposed structure |
| --- |
| 1. Obtain features with a feature importance score greater than zero using an embedded XGBoost method; |
| 2. Randomly initialize the population of each chromosome; |
| 3. do |

| 4.  | do |
| 5.  | Calculate the fitness value of each chromosome based on the fitness function; |
| 6.  | Select individuals from the population using the tournament selection method; |
| 7.  | Perform crossover between individuals using the uniform crossover method to generate offspring; |
| 8.  | Perform mutation between individuals using the simple mutation method to generate offspring; |
| 9.  | until the termination condition is satisfied |
| 10. | until the best performance |
| 11. | return the optimal gene subset; |
| 12. | divide into training and test sets based on 10-fold cross-validation; |
| 13. | return classification performance with different classifiers; |

Table 1. parameters of XGBoost

| Parameters | Value |
| --- | --- |
| n_estimators | 100 |
| max_depth | 3 |
| subsample | 0.75 |
| learning_rate | 0.3 |

Table 2. parameters of MOGA

| Parameters | Value |
| --- | --- |
| population | 100 |
| chromosome | initial features |
| fitness | K-nearest neighbor K=5 |
| iteration | 50 |
| crossover probability | 0.8 |
| mutation probability | 0.01 |

Next, we discuss the computational complexity of XGBoost-MOGA. The time complexity of XGBoost-MOGA depends on two stages. In the first stage, an XGBoost learning algorithm is used for the initial gene selection. The time complexity of this stage is $O(MlogN)$, where $M$ is the number of samples, and $N$ is the number of genes. In the second stage, genetic optimization is performed to search for the optimal gene subset, and the time complexity of this stage is $O(GPN')$ in the wrapper evaluations, where $G$ is the number of iterations, $P$ is the population size, and $N'$ is the size of the gene subset obtained by the first stage. In the summer, the time complexity of XGBoost-MOGA was $O(MlogN) + O(GPN')$. For the microarray gene dataset, we obtained $N \gg M$ and $GP \gg M$ in our approach. Hence, the computational complexity of XGBoost-MOGA was approximately $O(GPN')$.

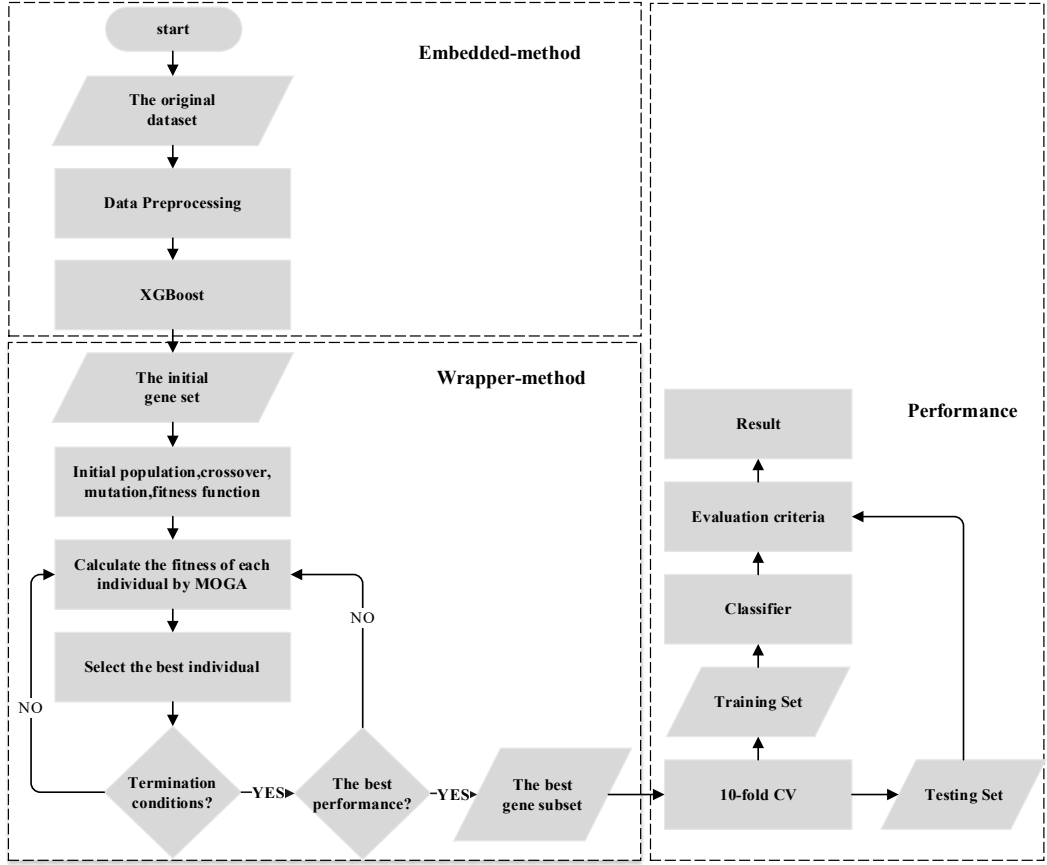

Fig. 2. Overall process of XGBoost-MOGA approach

## 4 Experimental results and analyses

We conducted experiments to verify the performance of the proposed approach. The datasets required for the experiment, classification algorithms, comparison of feature selection methods, and results and discussions are presented in this section. The results were generated on a PC equipped with a Core i5-5200U CPU and 12 G of memory.

4.1 Datasets

We performed our experiments on 13 publicly available microarray datasets. Table 3 outlines the characteristics of the 13 microarray datasets, including the numbers of genes, samples, and classes. Among these datasets, the number of genes ranged from 2000 to 20000, and the number of samples was less than 300. Only the Lymphoma_3 dataset was a three-class dataset, whereas the others were binary.

As some of the originally obtained datasets contained missing values, we performed missing value interpolation to process the missing values. In our experiments, we performed the k-nearest-neighbor interpolation to replace the missing values. Each sample's missing values were imputed using the mean value from n nearest neighbors in the training set. After the missing values were processed, the new standardized data were obtained [38]. We used the min–max normalization method to map the new data values to [0, 1].

Table 3. Microarray Datasets

| Dataset | No.genes | No.samples | No.classes |
| --- | --- | --- | --- |
| DLBCL [39] | 4026 | 47 | 2 |
| Breast [40] | 24481 | 97 | 2 |
| CNS [40] | 7128 | 60 | 2 |
| Colon [41] | 2000 | 62 | 2 |
| Leukemia [40] | 7129 | 72 | 2 |
| LungCancer [40] | 12533 | 181 | 2 |
| Lymphoma [39] | 7129 | 77 | 2 |
| N_A [42] | 10100 | 50 | 2 |
| Prostate [43] | 12600 | 102 | 2 |
| HD [44] | 22283 | 31 | 2 |
| Myeloma [45] | 12625 | 173 | 2 |
| Ovarian [40] | 15154 | 253 | 2 |
| Lymphoma_3 [46] | 4026 | 66 | 3 |

## 4.2 Experiment setting

For our experiments, we conducted two standard classifiers belonging to different paradigms, i.e., support vector machine (SVM) and naïve Bayes (NB), to evaluate the performance of selected genes for each gene selection method. The classification performances were recorded after a 10-fold cross-validation. All programs were implemented using the Python language. In particular, the these classification algorithms were implemented using the public package tools in XGBoost, scikit-feature [16], scikit-learn [47], and sklearn-genetic [48]. In our experiments, we implemented an SVM classifier with a linear kernel function. Other parameters involved in the SVM, and NB were set to the default values.

After completing the model construction, the performance of the model must be evaluated. When comparing the capabilities of different models, the use of different performance metrics will often yield different evaluation results. The simplest and most typically used criterion for evaluating a model is accuracy. However, if accuracy is used as an evaluation criterion without any premise, it often fails to reflect a model's performance. Hence, we used three additional evaluation criteria: F-score, precision, and recall for a comprehensive evaluation [8, 10, 49, 50]. Table 4 shows a binary classification confusion matrix [7], in which

True positive (TP): both the actual and predicted classes are positive examples;

False positive (FP): the actual and predicted classes are negative and positive examples, respectively;

False negative (FN): the actual and predicted classes are positive and negative examples, respectively;

True negative (TN): both the actual and predicted classes are negative examples.

Table 4. Confusion Matrix

| Confusion Matrix | | Predicted | |
|---|---|---|---|
| | | Positive | Negative |
| Actual | Positive | TP | FN |
| | Negative | FP | TN |

$$accuracy = \frac{TP + TN}{TP + FP + TN + FN} \quad (9)$$

$$P = precision = \frac{TP}{TP + FP} \quad (10)$$

$$R = recall = \frac{TP}{TP + FN} \quad (11)$$

$$F - score = \frac{2 \times P \times R}{P + R} \quad (12)$$

The calculation equations for accuracy, precision, recall, and F-score are shown in (9), (10), (11), and (12), respectively. In addition to accuracy, the other three evaluation criteria were for a single class. Therefore, we assigned the same weight to each class in multiclassification problems to calculate the three evaluation criteria values and their average values to obtain the final result.

4.3 Comparison of classification performance

In this section, we first compare the XGBoost-MOGA approach with its related methods, MOGA and XGBoost, to verify the effectiveness of XGBoost-MOGA. Subsequently, three other state-of-the-art feature selection methods are selected for comparison to demonstrate the advantages of XGBoost-MOGA.

Table 5 presents the numbers of genes selected by the MOGA, XGBoost, and XGBoost-MOGA. In addition, the numbers of genes in the original datasets are provided for reference. It was discovered that the number of genes selected by the MOGA was less than half of that in the original datasets. XGBoost selected a few genes from the original datasets. XGBoost-MOGA selected the least number of genes, i.e., less than or equal to half of that selected by XGBoost, in general.

Table 5. Number of selected genes by ORIGINAL, MOGA, XGBoost, and XGBoost-MOGA

| Dataset | ORIGINAL | MOGA | XGBoost | XGBoost-MOGA |
| --- | --- | --- | --- | --- |
| DLBCL | 4026 | 1943 | 32 | 12 |
| Breast | 24481 | 12231 | 137 | 28 |
| CNS | 7128 | 3481 | 97 | 56 |
| Colon | 2000 | 1010 | 76 | 62 |
| Leukemia | 7129 | 3552 | 13 | 7 |
| LungCancer | 12533 | 6222 | 23 | 5 |
| Lymphoma | 7129 | 3603 | 50 | 25 |
| N_A | 10100 | 4957 | 68 | 52 |
| Prostate | 12600 | 6251 | 64 | 54 |
| HD | 22283 | 10974 | 15 | 4 |
| Myeloma | 12625 | 6331 | 165 | 39 |
| Ovarian | 15154 | 7455 | 32 | 7 |
| Lymphoma_3 | 4026 | 1958 | 33 | 5 |

To assess the performances of the classification algorithms, the classifiers of SVM, and Naïve Bayes were used. We performed ten rounds of 10-fold cross-validation using the two classifiers on 13 selected genes set to obtain the average result. The optimal values obtained for each dataset are highlighted in bold.

Tables 6-7 show the average performances on the 13 datasets for the two classifiers based on four evaluation criteria. Compared with the other three methods, the proposed XGBoost-MOGA approach yielded better results in terms of accuracy, F-score, precision, and recall, in most cases. The proportions of XGBoost-MOGA's optimal accuracy were 12/13, and 11/13 for the abovementioned classifiers, respectively. In particular, XGBoost-MOGA achieved the best performance on the SVM classifier. As shown in Table 6, XGBoost-MOGA achieved 100% on seven datasets based on four evaluation criteria. For the XGBoost method, five datasets achieved 100% classification performance, as did five of the seven datasets of XGBoost-

MOGA. The performances of datasets of Ovarian and Lymphoma_3 reached 100% in the four methods based on the SVM classifier. As shown in Table 7, XGBoost and XGBoost-MOGA achieved 100% classification performance on four and three datasets, respectively. Among those datasets, three datasets were the same. The results of these four evaluation criteria indicate that XGBoost-MOGA performed the best, thereby verifying the effectiveness of XGBoost-MOGA.

Next, we compare XGBoost-MOGA with three other state-of-the-art feature selection methods: CFS [51], FCSVM-RFE [18], and MultiSURF [52]. The parameters of these three feature selection methods were set to default values. In order to compare with XGBoost-MOGA more fairly, except for CFS, FCSVM-RFE and MultiSURF select the same number of features as XGBoost-MOGA.

Furthermore, we performed ten rounds of 10-fold cross-validation to obtain the average value. Tables 8-9 show the accuracies of the two classifiers based on the four gene selection methods. The proportions of XGBoost-MOGA's optimal accuracy were 13/13, and 9/13, respectively, for the two classifiers mentioned earlier. Among the two classifiers, XGBoost-MOGA achieved the best performance using the SVM classifier. From the overall average classification accuracy, the classification accuracy of the SVM classifier is better than that of the NB classifier. Tables 8-9 show that none of the compared methods performed better than XGBoost-MOGA. The results suggest that XGBoost-MOGA for gene selection can substantially improve classification performance.

These comparisons show that the proposed XGBoost-MOGA outperforms all the state-of-the-art methods for 13 datasets. Therefore, based on our comprehensive evaluation, we conclude that our XGBoost-MOGA approach imposed a significant effect.

Table 6. Performances of SVM based on selected genes by ORIGINAL, MOGA, XGBoost, and XGBoost-MOGA

| Dataset | measurement | ORIGINAL | MOGA | XGBoost | XGBoost-MOGA |
|---|---|---|---|---|---|
| DLBCL | accuracy | **100.00** | 95.50 | 98.00 | **100.00** |
|  | F-score | **100.00** | 95.33 | 98.00 | **100.00** |
|  | precision | **100.00** | 96.67 | 98.33 | **100.00** |
|  | recall | **100.00** | 95.83 | 98.33 | **100.00** |
| Breast | accuracy | 71.00 | 71.11 | **83.67** | 82.33 |
|  | F-score | 70.72 | 70.54 | **83.53** | 81.82 |
|  | precision | 71.92 | 73.10 | **84.58** | 83.56 |
|  | recall | 71.17 | 71.33 | **83.75** | 82.08 |
| CNS | accuracy | 68.33 | 65.00 | 90.00 | **91.67** |
|  | F-score | 59.38 | 57.00 | 88.70 | **90.09** |
|  | precision | 59.92 | 57.92 | 91.33 | **92.58** |
|  | recall | 62.50 | 60.00 | **90.00** | **90.00** |
| Colon | accuracy | 85.71 | 87.14 | **90.24** | **90.24** |
|  | F-score | 83.83 | 85.27 | **89.04** | **89.04** |
|  | precision | 87.58 | 88.92 | **90.92** | **90.92** |
|  | recall | 85.00 | 85.42 | **89.58** | **89.58** |
| Leukemia | accuracy | 85.71 | **98.57** | 97.14 | **98.57** |
|  | F-score | 83.83 | **98.44** | 97.02 | **98.44** |
|  | precision | 87.58 | **99.00** | 97.75 | **99.00** |
|  | recall | 85.00 | **98.33** | 97.08 | **98.33** |
| LungCancer | accuracy | 98.57 | 99.44 | 98.89 | **100.00** |
|  | F-score | 98.44 | 98.84 | 97.19 | **100.00** |
|  | precision | 99.00 | 99.69 | 99.41 | **100.00** |
|  | recall | 98.33 | 98.33 | 96.67 | **100.00** |
| Lymphoma | accuracy | 95.89 | 95.89 | **100.00** | **100.00** |
|  | F-score | 94.30 | 94.30 | **100.00** | **100.00** |
|  | precision | 96.67 | 96.67 | **100.00** | **100.00** |
|  | recall | 94.17 | 94.17 | **100.00** | **100.00** |
| N_A | accuracy | 70.00 | 70.00 | **100.00** | **100.00** |
|  | F-score | 59.52 | 59.52 | **100.00** | **100.00** |
|  | precision | 57.75 | 57.75 | **100.00** | **100.00** |
|  | recall | 62.92 | 62.92 | **100.00** | **100.00** |
| Prostate | accuracy | 92.09 | 90.18 | 96.00 | **98.00** |
|  | F-score | 92.02 | 90.09 | 95.96 | **97.98** |
|  | precision | 92.92 | 91.49 | 96.67 | **98.33** |
|  | recall | 92.17 | 90.33 | 96.00 | **98.00** |
| HD | accuracy | 96.67 | 96.67 | **100.00** | **100.00** |
|  | F-score | 96.67 | 96.67 | **100.00** | **100.00** |
|  | precision | 97.50 | 97.50 | **100.00** | **100.00** |
|  | recall | 97.50 | 97.50 | **100.00** | **100.00** |

Table 6 (continued)

| Dataset | measurement | | | | |
|---|---|---|---|---|---|
| Myeloma | accuracy | 82.09 | 80.39 | 88.46 | **88.53** |
| | F-score | 63.59 | 61.06 | **81.49** | 79.23 |
| | precision | 76.06 | 69.81 | 84.26 | **89.85** |
| | recall | 62.77 | 60.39 | **80.84** | 76.46 |
| Ovarian | accuracy | **100.00** | **100.00** | **100.00** | **100.00** |
| | F-score | **100.00** | **100.00** | **100.00** | **100.00** |
| | precision | **100.00** | **100.00** | **100.00** | **100.00** |
| | recall | **100.00** | **100.00** | **100.00** | **100.00** |
| Lymphoma_3 | accuracy | **100.00** | **100.00** | **100.00** | **100.00** |
| | F-score | **100.00** | **100.00** | **100.00** | **100.00** |
| | precision | **100.00** | **100.00** | **100.00** | **100.00** |
| | recall | **100.00** | **100.00** | **100.00** | **100.00** |
| Winners | accuracy | 3 | 3 | 8 | **12** |
| | F-score | 3 | 3 | 8 | **12** |
| | precision | 3 | 3 | 8 | **12** |
| | recall | 3 | 3 | 8 | **12** |

Table 7. Performances of NB based on selected genes by ORIGINAL, MOGA, XGBoost, and XGBoost-MOGA

| Dataset | measurement | ORIGINAL | MOGA | XGBoost | XGBoost-MOGA |
|---|---|---|---|---|---|
| DLBCL | accuracy | 96.00 | 92.00 | **100.00** | **100.00** |
| | F-score | 95.62 | 91.83 | **100.00** | **100.00** |
| | precision | 97.08 | 92.50 | **100.00** | **100.00** |
| | recall | 95.83 | 92.50 | **100.00** | **100.00** |
| Breast | accuracy | 48.67 | 46.67 | 54.67 | **56.67** |
| | F-score | 41.50 | 40.63 | 41.04 | **42.44** |
| | precision | **44.46** | 42.93 | 42.01 | 40.31 |
| | recall | 47.58 | 45.58 | 52.25 | **54.42** |
| CNS | accuracy | 70.00 | 65.00 | **83.33** | 83.33 |
| | F-score | 65.69 | 60.13 | **76.88** | 76.06 |
| | precision | 69.50 | 61.17 | 76.92 | **77.17** |
| | recall | 65.00 | 60.42 | **78.75** | 77.50 |
| Colon | accuracy | 58.10 | 59.29 | **77.14** | **77.14** |
| | F-score | 56.29 | 58.15 | 75.46 | **75.87** |
| | precision | 59.42 | 60.17 | 80.67 | **81.92** |
| | recall | 62.92 | 62.08 | 78.75 | **80.00** |
| Leukemia | accuracy | **98.57** | 97.14 | 97.14 | **98.57** |
| | F-score | **98.44** | 97.02 | 97.02 | **98.44** |
| | precision | **99.00** | 97.75 | 97.75 | **99.00** |
| | recall | **98.33** | 97.08 | 97.08 | **98.33** |
| LungCancer | accuracy | 98.33 | 99.44 | **100.00** | 98.89 |
| | F-score | 96.30 | 99.11 | **100.00** | 98.23 |
| | precision | 98.16 | 98.75 | **100.00** | 97.50 |
| | recall | 96.33 | 99.67 | **100.00** | 99.33 |

Table 7 (continued)

| | | | | | |
|---|---|---|---|---|---|
| Lymphoma | accuracy | 80.71 | 79.29 | 96.07 | **97.32** |
| | F-score | 70.59 | 68.22 | 94.97 | **96.42** |
| | precision | 73.06 | 68.94 | 95.83 | **97.50** |
| | recall | 72.50 | 71.67 | 95.83 | **96.67** |
| N_A | accuracy | 70.00 | 70.00 | 94.00 | **96.00** |
| | F-score | 54.81 | 57.44 | 89.68 | **95.62** |
| | precision | 52.67 | 54.83 | 90.25 | **97.08** |
| | recall | 60.42 | 61.67 | 91.25 | **95.83** |
| Prostate | accuracy | 62.73 | 62.73 | **93.09** | **93.09** |
| | F-score | 58.15 | 58.15 | **93.03** | **93.03** |
| | precision | 64.52 | 64.52 | **93.75** | **93.75** |
| | recall | 62.50 | 62.50 | **93.17** | **93.17** |
| HD | accuracy | 84.17 | 84.17 | **100.00** | **100.00** |
| | F-score | 84.00 | 84.00 | **100.00** | **100.00** |
| | precision | 88.33 | 88.33 | **100.00** | **100.00** |
| | recall | 87.50 | 87.50 | **100.00** | **100.00** |
| Myeloma | accuracy | 79.22 | 76.34 | 80.42 | **81.54** |
| | F-score | 60.44 | 58.90 | 69.11 | **74.24** |
| | precision | 65.30 | 61.86 | 69.67 | **73.95** |
| | recall | 61.46 | 59.62 | 72.24 | **77.84** |
| Ovarian | accuracy | 91.31 | 90.91 | **99.22** | **99.22** |
| | F-score | 90.58 | 90.11 | **99.14** | 99.12 |
| | precision | 90.42 | 90.02 | 99.21 | **99.43** |
| | recall | 91.07 | 90.51 | **99.15** | 98.89 |
| Lymphoma_3 | accuracy | 95.48 | 95.48 | **100.00** | **100.00** |
| | F-score | 89.02 | 89.02 | **100.00** | **100.00** |
| | precision | 88.22 | 88.22 | **100.00** | **100.00** |
| | recall | 90.00 | 90.00 | **100.00** | **100.00** |
| Winners | accuracy | 1 | 0 | 8 | **11** |
| | F-score | 1 | 0 | 7 | **10** |
| | precision | 2 | 0 | 6 | **10** |
| | recall | 1 | 0 | 7 | **11** |

Table 8. Accuracy of SVM based on selected genes by XGBoost-MOGA,CFS, FCSVM-RFE, and MultiSURF

| Dataset | XGBoost-MOGA | CFS | FCSVM-RFE | MultiSURF |
|---|---|---|---|---|
| DLBCL | **100.00 (+/- 0.00)** | 67.50 (+/- 26.58) | 100.00 (+/- 0.00) | 96.00 (+/- 8.00) |
| Breast | **82.33 (+/- 14.15)** | 52.56 (+/- 4.77) | 59.00 (+/- 13.50) | 66.67 (+/- 22.15) |
| CNS | **91.67 (+/- 11.18)** | 68.33 (+/- 8.98) | 58.33 (+/- 11.18) | 83.33 (+/- 12.91) |
| Colon | **90.24 (+/- 10.93)** | 64.76 (+/- 3.81) | 83.81 (+/- 12.95) | 85.24 (+/- 11.65) |
| Leukemia | **98.57 (+/- 4.29)** | 76.25 (+/- 19.27) | 79.11 (+/- 9.52) | 94.29 (+/- 7.00) |
| LungCancer | **100.00 (+/- 0.00)** | 89.53 (+/- 3.76) | 87.89 (+/- 5.24) | 97.78 (+/- 3.69) |
| Lymphoma | **100.00 (+/- 0.00)** | 75.36 (+/- 3.73) | 83.04 (+/- 8.31) | 97.32 (+/- 5.37) |

Table 8 (continued)

| | | | | |
|---|---|---|---|---|
| N_A | **100.00 (+/- 0.00)** | 66.00 (+/- 9.17) | 64.00 (+/- 12.00) | 82.00 (+/- 10.77) |
| Prostate | **98.00 (+/- 4.00)** | 89.27 (+/- 8.14) | 78.27 (+/- 14.12) | 94.09 (+/- 6.58) |
| HD | **100.00 (+/- 0.00)** | 70.83 (+/- 23.35) | 96.67 (+/- 10.00) | 96.67 (+/- 10.00) |
| Myeloma | **88.53 (+/- 4.94)** | 79.22 (+/- 2.61) | 79.22 (+/- 2.61) | 84.31 (+/- 8.40) |
| Ovarian | **100.00 (+/- 0.00)** | 99.20 (+/- 1.60) | 95.29 (+/- 3.81) | 96.85 (+/- 2.94) |
| Lymphoma_3 | **100.00 (+/- 0.00)** | 90.71 (+/- 7.64) | 98.57 (+/- 4.29) | 77.38 (+/- 7.08) |
| Mean | **96.10** | 76.12 | 81.78 | 88.61 |

Table 9. Accuracy of NB based on selected genes by XGBoost-MOGA,CFS, FCSVM-RFE, and MultiSURF

| Dataset | XGBoost-MOGA | CFS | FCSVM-RFE | MultiSURF |
|---|---|---|---|---|
| DLBCL | **100.00 (+/- 0.00)** | 77.00 (+/- 22.72) | 98.00 (+/- 6.00) | 94.00 (+/- 9.17) |
| Breast | 56.67 (+/- 8.65) | 53.67 (+/- 3.98) | 56.56 (+/- 13.65) | **59.67 (+/- 14.16)** |
| CNS | **83.33 (+/- 14.91)** | 60.00 (+/- 17.00) | 53.33 (+/- 23.33) | 81.67 (+/- 15.72) |
| Colon | 77.14 (+/- 15.03) | 45.00 (+/- 19.30) | 52.86 (+/- 18.62) | **85.71 (+/- 10.96)** |
| Leukemia | **98.57 (+/- 4.29)** | 75.89 (+/- 21.33) | 93.21 (+/- 8.81) | 97.14 (+/- 5.71) |
| LungCancer | **98.89 (+/- 2.22)** | 96.70 (+/- 2.70) | 95.00 (+/- 3.89) | 97.84 (+/- 4.33) |
| Lymphoma | **97.32 (+/- 5.37)** | 72.86 (+/- 12.00) | 71.25 (+/- 14.95) | 89.64 (+/- 12.72) |
| N_A | **96.00 (+/- 8.00)** | 68.00 (+/- 16.00) | 54.00 (+/- 20.10) | 94.00 (+/- 9.17) |
| Prostate | 93.09 (+/- 8.97) | 88.18 (+/- 7.54) | 62.82 (+/- 15.67) | **94.09 (+/- 6.58)** |
| HD | **100.00 (+/- 0.00)** | 78.33 (+/- 23.63) | 94.17 (+/- 11.81) | 96.67 (+/- 10.00) |
| Myeloma | 81.54 (+/- 6.18) | 70.52 (+/- 8.17) | 64.31 (+/- 11.72) | **82.06 (+/- 7.17)** |
| Ovarian | **99.22 (+/- 1.57)** | 80.63 (+/- 5.97) | 86.55 (+/- 7.75) | 96.85 (+/- 2.94) |
| Lymphoma_3 | **100.00 (+/- 0.00)** | 89.52 (+/- 9.42) | 98.57 (+/- 4.29) | 82.14 (+/- 15.40) |
| Mean | **90.91** | 73.56 | 75.43 | 88.58 |

## 4.4 Statistical analysis

We used the Wilcoxon signed-rank test [53, 54] to evaluate whether the differences between XGBoost-MOGA and the compared gene selection methods were statistically significant based on 13 publicly available microarray datasets. The Wilcoxon signed-rank test is a nonparametric hypothesis testing method that is generally used to test whether a difference exists between two paired samples' data. In the experiment, we set the Wilcoxon signed-rank test significance level value to 0.05 [2]. Statistical significance was defined as a p-value of $< 0.05$ [55, 56]. We used Python's Wilcoxon signed-rank test algorithm package for the rank test. Tables 10 and 11 show the matrices obtained by the Wilcoxon signed-rank test XGBoost-MOGA and the compared gene selection methods on the two classifiers, respectively. The values in the two tables show the corresponding p-values.

Table 10 shows the p-values of the accuracy test results of XGBoost-MOGA vs. its related methods using two classifiers. The table contained 6 p-values, and each p-value corresponded to a pair of comparisons. XGBoost-MOGA vs. all compared methods was statistically different, but only XGBoost-MOGA could not statistically outperform XGBoost gene selection method for SVM classifier. In other words, our proposed gene selection method outperformed 5 times among 6 pairs of comparisons.

Table 11 shows the p-values of the accuracy significance test results of XGBoost-MOGA vs. three other state-of-the-art feature selection methods using two classifiers. The table contained 6 p-values, and each p-value corresponded to a pair of comparisons. It was observed that XGBoost-MOGA vs. all compared methods was statistically different, but only XGBoost-MOGA could not statistically outperform MultiSURF for NB classifiers. This means that our proposed gene selection approach outperformed 5 times among 6 pairs of comparisons.

In summary, the analysis of the two sets of experimental results shows statistical difference between XGBoost-MOGA and the compared gene selection methods. Additionally, this verifies that XGBoost-MOGA is extremely effective.

Table 10. Wilcoxon signed-rank test matrix for XGBoost-MOGA with ORIGINAL, MOGA, and XGBoost based on two classifiers

|  | XGBoost-MOGA | |
|---|---|---|
|  | SVM | NB |
| ORIGINAL | 0.005062 | 0.002218 |
| MOGA | 0.005062 | 0.000488 |
| XGBoost | 0.062509 | 0.045800 |

Table 11. Wilcoxon signed-rank test matrix for XGBoost-MOGA with three other state-of-the-art feature selection methods based on two classifiers

|  | XGBoost-MOGA | |
|---|---|---|
|  | SVM | NB |
| CFS | 0.000244 | 0.000244 |
| FCSVM-RFE | 0.002218 | 0.000244 |
| MultiSURF | 0.000244 | 0.127197 |

4.5 Comparison of complexity

In this section, we analyze the computational complexity of the gene selection algorithms used in the experiment.

Table 12 lists the time complexity of all gene selection algorithms compared in our experiment. In this table, $M$ is the number of samples, $N$ is the number of genes, $G$ is the number of iterations, $P$ is the population size, $N'$ is the size of the gene subset obtained by the first stage, $K$ is the required number of clusters, and $T$ is the number of iterations.

Table 12. Comparisons of the computational complexity of all compared gene selection algorithms

| Algorithms | Complexity |
| --- | --- |
| MOGA | $O(GPN)$ |
| XGBoost | $O(MlogN)$ |
| CFS | $O(MN^2)$ |
| FCSVM-RFE | $O(MNKT)$ |
| MultiSURF | $O(NM^2)$ |
| XGBoost-MOGA | $O(GPN')$ |

Table 13 lists the runtime of gene selection for the MOGA, XGBoost, and XGBoost-MOGA on each dataset. The numbers are reported as the average of 10 rounds of execution. The three methods can be directly compared by verifying the mean values in the last row in Table 13. As shown, XGBoost indicated the shortest execution time; MOGA required the longest time; XGBoost-MOGA required a longer time than XGBoost but less time than the MOGA. Although XGBoost indicated the shortest execution time, it was significantly worse than XGBoost-MOGA in terms of accuracy, F-score, precision, recall, and the number of selected genes.

Table 14 shows the execution time of gene selection for XGBoost-MOGA, CFS, FCSVM-RFE, and MultiSURF on each dataset. As shown, the execution time of XGBoost-MOGA was much shorter than those of CFS, and MultiSURF. Although FCSVM-RFE required the least mean execution time, it ranks third in terms of accuracy. CFS required the longest execution time.

As shown, the proposed approach's mean average execution time was better than those of most gene selection methods compared, and it performed significantly better than all the gene selection methods compared. From our comprehensive evaluation, it was evident that XGBoost-MOGA imposed a significant effect.

Table 13. Comparison of runtime (in s) of gene selection for MOGA, XGBoost, and XGBoost-MOGA

| Dataset | MOGA | XGBoost | XGBoost-MOGA |
| --- | --- | --- | --- |
| DLBCL | 473.9519 | 1.5469 | 243.5371 |
| Breast | 8027.6693 | 34.2477 | 487.1160 |
| CNS | 837.1951 | 7.4949 | 227.6366 |
| Colon | 341.9173 | 1.2400 | 202.3151 |
| Leukemia | 1890.9567 | 3.4406 | 290.9553 |
| LungCancer | 11047.1821 | 11.5957 | 417.5358 |

Table 13 (continued)

|  |  |  |  |
|---|---|---|---|
| Lymphoma | 2356.3472 | 5.2414 | 228.3122 |
| N_A | 1252.2147 | 6.4353 | 283.0818 |
| Prostate | 5257.5008 | 9.4688 | 343.2004 |
| HD | 1388.1430 | 7.5244 | 260.9775 |
| Myeloma | 14610.2058 | 25.5567 | 435.4780 |
| Ovarian | 21497.0954 | 23.5234 | 409.1696 |
| Lymphoma_3 | 887.4904 | 5.4856 | 229.8213 |
| Mean | 5374.4515 | 10.9847 | 312.2413 |

Table 14. Comparison of runtime (in s) of gene selection for XGBoost-MOGA, CFS, FCSVM-RFE, and MultiSURF

| Dataset | XGBoost-MOGA | CFS | FCSVM-RFE | MultiSURF |
|---|---|---|---|---|
| DLBCL | 243.5371 | 173.0191 | 2.8394 | 24.3121 |
| Breast | 487.1160 | 4334.6192 | 23.8893 | 45.9477 |
| CNS | 227.6366 | 16952.1852 | 4.6883 | 91.6645 |
| Colon | 202.3151 | 102.6703 | 1.0804 | 29.9868 |
| Leukemia | 290.9553 | 1082.2436 | 5.1461 | 127.2141 |
| LungCancer | 417.5358 | 3836.1710 | 14.0079 | 843.9765 |
| Lymphoma | 228.3122 | 1125.2366 | 5.0691 | 116.3613 |
| N_A | 283.0818 | 850.7205 | 6.4203 | 86.3056 |
| Prostate | 343.2004 | 881.9791 | 14.0729 | 545.9592 |
| HD | 260.9775 | 1047.4465 | 9.8434 | 94.3040 |
| Myeloma | 435.4780 | 1496.2270 | 16.2907 | 1029.6611 |
| Ovarian | 409.1696 | 13825.5552 | 18.2016 | 1947.8391 |
| Lymphoma_3 | 229.8213 | 630.6626 | 2.8484 | 87.8956 |
| Mean | 312.2413 | 3564.5181 | 9.5691 | 390.1098 |

4.6 Gene analysis

The selection of genes (i.e., biomarkers) associated with cancer is crucial. In order to demonstrate the utility of the proposed approach in identifying biomarkers, we provide biological descriptions of the selected genes on four microarray datasets. Tables 15-18 list the genes selected by XGBoost-MOGA on four microarray datasets. These four tables involve the selected gene index, Probe/uniprot ID, Ensembl ID, Gene, and the description of the gene, respectively. In view of the limited space, we investigated the functions of the genes selected in the Leukemia dataset and described below.

MCFD2 encodes a soluble luminal protein with two calmodulin-like EF-hand motifs at its C-terminus. This gene protein has also been shown to maintain stem cell potential in adult central nervous system and is a marker for testicular germ cell tumors.

ALOX5AP, together with 5-lipoxygenase, encodes a protein that is necessary for the synthesis of leukotrienes. The protein is located in the plasma membrane. Its functional inhibitor prevents the transport of 5-lipoxygenase from the cytoplasm to the cell membrane and inhibits the activation of 5-lipoxygenase.

CD59 is a gene that encodes a cell surface glycoprotein, which regulates complement-mediated cell lysis and involves in lymphocyte signal transduction. Mutations in this gene can cause CD59 deficiency, a disease that causes hemolytic anemia and thrombosis, and can lead to cerebral infarction.

UBE2S also known as EPF5, E2EPF, and E2-EPF, is a member of the ubiquitin-conjugating enzyme family. The encoded protein can form a thiol ester linkage with ubiquitin in an ubiquitin-activating enzyme-dependent manner, which is a characteristic property of the ubiquitin carrier protein.

ABCC6 is a member of the superfamily of ATP-binding cassette (ABC) transporters. The protein encoded by this gene is involved in multi-drug resistance. Mutations in this gene can cause pseudoxanthoma elasticum.

ATM belongs to the PI3/pi4 kinase family that encodes a protein. Therefore, it acts as a regulatory protein for a variety of downstream proteins, including tumor suppressor proteins p53 and BRCA1, checkpoint kinase CHK2, checkpoint proteins RAD17 and RAD9, and DNA repair proteins NBS1. Mutations in this gene are related to ataxia telangiectasia, which is an autosomal recessive genetic disease.

TERF1 gene encodes a telomere-specific protein, which is part of the telomere nuclear protein complex. This protein exists on telomeres throughout the cell cycle and acts as an inhibitor of telomerase, limiting the extension of the ends of a single chromosome in cis.

Table 15 Descriptions of the Leukemia genes selected by XGBoost-MOGA

| Index | Probe/uniprot ID | Ensembl ID | Gene | Description |
|---|---|---|---|---|
| 1833 | M23161_AT | ENSG00000180398 | MCFD2 | multiple coagulation factor deficiency 2, ER cargo receptor complex subunit |
| 2127 | M63262_AT | ENSG00000132965 | ALOX5AP | arachidonate 5-lipoxygenase activating protein |
| 2287 | M84349_AT | ENSG00000085063 | CD59 | CD59 molecule (CD59 blood group) |
| 2353 | M91670_AT | ENSG00000108106 | UBE2S | ubiquitin conjugating enzyme E2 S |
| 4846 | X95715_AT | ENSG00000091262 | ABCC6 | ATP binding cassette subfamily C member 6 |
| 6538 | X91196_S_AT | ENSG00000149311 | ATM | ATM serine/threonine kinase |
| 6854 | U74382_S_AT | ENSG00000147601 | TERF1 | telomeric repeat binding factor 1 |

Table 16 Descriptions of the LungCancer genes selected by XGBoost-MOGA

| Index | Probe/uniprot ID | Ensembl ID | Gene | Description |
|---|---|---|---|---|
| 3360 | 34319_AT | ENSG00000163993 | S100P | S100 calcium binding protein P |
| 3915 | 33354_AT | ENSG00000108854 | SMURF2 | SMAD specific E3 ubiquitin protein ligase 2 |
| 4335 | 33903_AT | ENSG00000167657 | DAPK3 | death associated protein kinase 3 |
| 7624 | 37577_AT | ENSG00000213390 | ARHGAP19 | Rho GTPase activating protein 19 |
| 8536 | 38481_AT | ENSG00000132383 | RPA1 | replication protein A1 |

Table 17 Descriptions of the Breast genes selected by XGBoost-MOGA

| Index | Probe/uniprot ID | Ensembl ID | Gene | Description |
|---|---|---|---|---|
| 156 | AB033049 | ENSG00000151458 | ANKRD50 | ankyrin repeat domain 50 |
| 393 | Contig1815_RC | | | |
| 589 | NM_002376 | ENSG00000075413 | MARK3 | microtubule affinity regulating kinase 3 |
| 718 | NM_000956 | ENSG00000125384 | PTGER2 | prostaglandin E receptor 2 |
| 1408 | NM_002485 | ENSG00000104320 | NBN | nibrin |
| 1683 | Contig13506_RC | | | |
| 3189 | Contig11133_RC | | | |
| 3696 | NM_002748 | ENSG00000069956 | MAPK6 | mitogen-activated protein kinase 6 |
| 4632 | NM_002890 | ENSG00000145715 | RASA1 | RAS p21 protein activator 1 |
| 5205 | NM_005051 | ENSG00000172053 | QARS1 | glutaminyl-tRNA synthetase 1 |
| 5622 | NM_002944 | ENSG00000047936 | ROS1 | ROS proto-oncogene 1, receptor tyrosine kinase |
| 5936 | Contig50360_RC | | | |
| 7434 | NM_003854 | ENSG00000115598 | IL1RL2 | interleukin 1 receptor like 2 |
| 8661 | Contig42342_RC | | | |
| 9030 | NM_005443 | ENSG00000138801 | PAPSS1 | 3'-phosphoadenosine 5'-phosphosulfate synthase 1 |
| 9938 | Contig56089_RC | | | |
| 10826 | NM_004993 | ENSG00000066427 | ATXN3 | ataxin 3 |
| 11412 | NM_007178 | ENSG00000023734 | STRAP | serine/threonine kinase receptor associated protein |
| 11739 | AF038202 | ENSG00000135823 | STX6 | syntaxin 6 |
| 12221 | L40391 | ENSG00000170348 | TMED10 | transmembrane p24 trafficking protein 10 |
| 12827 | Contig10686_RC | | | |
| 13308 | NM_006680 | ENSG00000151376 | ME3 | malic enzyme 3 |
| 14446 | Contig9042_RC | | | |
| 17704 | Contig39797_RC | | | |
| 17897 | Contig4380_RC | | | |
| 21873 | AL050172 | ENSG00000173914 | RBM4B | RNA binding motif protein 4B |
| 22612 | Contig34329_RC | | | |
| 22859 | NM_001343 | ENSG00000153071 | DAB2 | DAB adaptor protein 2 |

Table 18 Descriptions of the Lymphoma_3 genes selected by XGBoost-MOGA

| Index | Probe/uniprot ID | Ensembl ID | Gene | Description |
|---|---|---|---|---|
| 677 | GENE2295X | ENSG00000137273 | FOXF2 | forkhead box F2 |
| 755 | GENE2398X | | FRA5E | fragile site, aphidicolin type, common, fra(5)(p14) |
| 1006 | GENE2110X | ENSG00000171503 | ETFDH | electron transfer flavoprotein dehydrogenase |
| 2732 | GENE621X | ENSG00000169071 | ROR2 | receptor tyrosine kinase like orphan receptor 2 |
| 2761 | GENE657X | ENSG00000107779 | BMPR1A | bone morphogenetic protein receptor type 1A |

# 5 Conclusion and future work

Gene selection is crucial in the analysis of microarray gene expression data. Selecting a minimal number of relevant genes from microarray data is crucial to cancer classification. Herein, we proposed a hybrid gene selection approach using XGBoost and a multi-objective genetic algorithm, namely XGBoost-MOGA, for cancer classification in microarray datasets. A series of experimental results analysis showed that XGBoost-MOGA offered significant advantages, i.e., it selected the least number of features and performed significantly better than other state-of-the-art gene selection methods.

In our experiment, XGBoost-MOGA involved two parts of parameters, among which the MOGA parameters were artificially set in advance. In future studies, we will investigate the implementation of adaptive parameter selection in the MOGA stage. In addition, we will attempt to incorporate some sampling methods to further improve classification performance by considering that the microarray datasets are primarily imbalanced.

**Acknowledgements**

Research on this work was partially supported by the grants from the National Science Foundation of China (No. 61562061), the funds from Jiangxi Education Department (No.GJJ1611096).

**References**

1. Güçkıran K, Cantürk İ, Özyılmaz L (2019) LASSO ve Relief Özellik Seçimi Yöntemleri ile DVM, ÇKA ve RO Ağ Yapıları Kullanılarak DNA Mikroçip Gen İfadesi Verisetlerinin Sınıflandırılması. Süleyman Demirel Üniversitesi Fen Bilimleri Enstitüsü Dergisi 23:115–121. https://doi.org/10.19113/sdufenbed.453462

2. Lazar C, Taminau J, Meganck S, et al (2012) A Survey on Filter Techniques for Feature Selection in Gene Expression Microarray Analysis. IEEE/ACM Trans Comput Biol and Bioinf 9:1106–1119. https://doi.org/10.1109/TCBB.2012.33

3. Lee C-P, Leu Y (2011) A novel hybrid feature selection method for microarray data analysis. Applied Soft Computing 11:208–213. https://doi.org/10.1016/j.asoc.2009.11.010

4. Hira ZM, Gillies DF (2015) A Review of Feature Selection and Feature Extraction Methods Applied on Microarray Data. Advances in Bioinformatics 2015:1–13. https://doi.org/10.1155/2015/198363

5. Bhalla A, Agrawal RK (2013) Microarray gene-expression data classification using less gene expressions by combining feature selection methods and classifiers. IJIEEB 5:42–48. https://doi.org/10.5815/ijieeb.2013.05.06

6. Bindu NH, Chakravarthi T (2018) Booster of an FS Algorithm on High Dimensional Data. IJSRSET 4:496–500

7. Yu H, Ni J (2014) An Improved Ensemble Learning Method for Classifying High-Dimensional and Imbalanced Biomedicine Data. IEEE/ACM Transactions on Computational Biology & Bioinformatics 11:657–666

8. Li M, Xiong A, Wang L, et al (2020) ACO Resampling: Enhancing the performance of oversampling methods for class imbalance classification. Knowledge-Based Systems 196:105818

9. Li W, Yin Y, Quan X, Zhang H (2019) Gene Expression Value Prediction Based on XGBoost Algorithm. Front Genet 10:1077-. https://doi.org/10.3389/fgene.2019.01077

10. Islam A, Rahman MM, Ahmed E, et al (2020) Adaptive Feature Selection and Classification of Colon Cancer From Gene Expression Data: an Ensemble Learning Approach. In: Proceedings of the International Conference on Computing Advancements. ACM, Dhaka Bangladesh, pp 1–7

11. Kavitha KR, Gopinath A, Gopi M (2017) Applying improved svm classifier for leukemia cancer classification using FCBF. In: 2017 International Conference on Advances in Computing, Communications and Informatics (ICACCI). pp 61–66

12. Ben Brahim A, Limam M (2013) Robust ensemble feature selection for high dimensional data sets. In: 2013 International Conference on High Performance Computing & Simulation (HPCS). IEEE, Helsinki, Finland, pp 151–157

13. Hall MA, Smith LA (1999) Feature Selection for Machine Learning: Comparing a Correlation-Based Filter Approach to the Wrapper. In: Proceedings of the Twelfth International Florida Artificial Intelligence Research Society Conference, May 1-5, 1999, Orlando, Florida, USA

14. Zeng X-Q, Li G-Z, Chen S-F (2010) Gene selection by using an improved Fast Correlation-Based Filter. In: 2010 IEEE International Conference on Bioinformatics and Biomedicine Workshops (BIBMW). IEEE, HongKong, China, pp 625–630

15. Chandrashekar G, Sahin F (2014) A survey on feature selection methods. Computers & Electrical Engineering 40:16–28. https://doi.org/10.1016/j.compeleceng.2013.11.024

16. Li J, Cheng K, Wang S, et al (2018) Feature Selection: A Data Perspective. ACM Comput Surv 50:1–45. https://doi.org/10.1145/3136625

17. Elyasigomari V, Lee DA, Screen HRC, Shaheed MH (2017) Development of a two-stage gene selection method that incorporates a novel hybrid approach using the cuckoo optimization algorithm and harmony search for cancer classification. Journal of Biomedical Informatics 67:11–20. https://doi.org/10.1016/j.jbi.2017.01.016

18. Huang X, Zhang L, Wang B, et al (2018) Feature clustering based support vector machine recursive feature elimination for gene selection. Appl Intell 48:594–607. https://doi.org/10.1007/s10489-017-0992-2

19. Shukla AK, Singh P, Vardhan M (2019) A New Hybrid Feature Subset Selection Framework Based on Binary Genetic Algorithm and Information Theory. Int J Comp Intel Appl 18:1950020. https://doi.org/10.1142/S1469026819500202

20. Huan Liu, Setiono R (1995) Chi2: feature selection and discretization of numeric attributes. In: Proceedings of 7th IEEE International Conference on Tools with Artificial Intelligence. IEEE Comput. Soc. Press, Herndon, VA, USA, pp 388–391

21. Liu Y (2004) A Comparative Study on Feature Selection Methods for Drug Discovery. J Chem Inf Comput Sci 44:1823–1828. https://doi.org/10.1021/ci049875d

22. Robnik-Šikonja M, Kononenko I (2003) Theoretical and Empirical Analysis of ReliefF and RReliefF. Machine Learning 53:23–69. https://doi.org/10.1023/A:1025667309714

23. Ghosh M, Adhikary S, Ghosh KK, et al (2019) Genetic algorithm based cancerous gene identification from microarray data using ensemble of filter methods. Med Biol Eng Comput 57:159–176. https://doi.org/10.1007/s11517-018-1874-4

24. Fleuret F (2004) Fast Binary Feature Selection with Conditional Mutual Information. J Mach Learn Res 5:1531–1555

25. Lu H, Chen J, Yan K, et al (2017) A hybrid feature selection algorithm for gene expression data classification. Neurocomputing 256:56–62. https://doi.org/10.1016/j.neucom.2016.07.080

26. Chen T, Guestrin C (2016) XGBoost: A Scalable Tree Boosting System. In: Proceedings of the 22nd ACM SIGKDD International Conference on Knowledge Discovery and Data Mining. ACM, San Francisco California USA, pp 785–794

27. Friedman JH (2001) Greedy Function Approximation: A Gradient Boosting Machine. The Annals of Statistics 29:1189–1232

28. Chen S, Zhou W, Tu J, et al (2021) A Novel XGBoost Method to Infer the Primary Lesion of 20 Solid Tumor Types From Gene Expression Data. Front Genet 12:632761. https://doi.org/10.3389/fgene.2021.632761

29. Islam A, Rahman MM, Ahmed E, et al (2020) Adaptive Feature Selection and Classification of Colon Cancer From Gene Expression Data: an Ensemble Learning Approach. In: Proceedings of the International Conference on Computing Advancements. Association for Computing Machinery, New York, NY, USA, pp 1–7

30. Dimitrakopoulos GN, Vrahatis AG, Plagianakos V, Sgarbas K (2018) Pathway analysis using XGBoost classification in Biomedical Data. In: Proceedings of the 10th Hellenic Conference on Artificial Intelligence. ACM, Patras Greece, pp 1–6

31. Sujamol S, Vimina ER, Krishnakumar U (2020) Improving Recurrence Prediction Accuracy of Ovarian Cancer Using Multi-phase Feature Selection Methodology. Applied Artificial Intelligence 35:1–21. https://doi.org/10.1080/08839514.2020.1854988
32. Abdu-Aljabar RD, Awad OA (2021) A Comparative analysis study of lung cancer detection and relapse prediction using XGBoost classifier. IOP Conf Ser: Mater Sci Eng 1076:012048. https://doi.org/10.1088/1757-899X/1076/1/012048
33. Haidar A, Verma B, Haidar R (2019) A Swarm based Optimization of the XGBoost Parameters. Aust J Intell Inf Process Syst 16:74–81
34. Djellali H, Guessoum S, Ghoualmi-Zine N, Layachi S (2017) Fast correlation based filter combined with genetic algorithm and particle swarm on feature selection. In: 2017 5th International Conference on Electrical Engineering - Boumerdes (ICEE-B). IEEE, Boumerdes, pp 1–6
35. Pragadeesh C, Jeyaraj R, Siranjeevi K, et al (2019) Hybrid feature selection using micro genetic algorithm on microarray gene expression data. IFS 36:2241–2246. https://doi.org/10.3233/JIFS-169935
36. Babatunde OH, Armstrong L, Leng J, Diepeveen D (2014) A Genetic Algorithm-Based Feature Selection. british journal of mathematics & computer science 5:889–905
37. Sayed S, Nassef M, Badr A, Farag I (2019) A Nested Genetic Algorithm for feature selection in high-dimensional cancer Microarray datasets. Expert Systems with Applications 121:233–243. https://doi.org/10.1016/j.eswa.2018.12.022
38. Song K, Yan F, Ding T, et al (2020) A steel property optimization model based on the XGBoost algorithm and improved PSO. Computational Materials Science 174:109472. https://doi.org/10.1016/j.commatsci.2019.109472
39. Alizadeh AA, Eisen MB, Davis RE, et al (2000) Distinct types of diffuse large B-cell lymphoma identified by gene expression profiling. Nature 403:503–511. https://doi.org/10.1038/35000501
40. Zhu Z, Ong Y-S, Dash M (2007) Markov blanket-embedded genetic algorithm for gene selection. Pattern Recognition 40:3236–3248. https://doi.org/10.1016/j.patcog.2007.02.007
41. Alon U, Barkai N, Notterman DA, et al (1999) Broad patterns of gene expression revealed by clustering analysis of tumor and normal colon tissues probed by oligonucleotide arrays. Proceedings of the National Academy of Sciences 96:6745–6750. https://doi.org/10.1073/pnas.96.12.6745
42. Subramanian AA, Tamayo PP, Mootha VKV, et al (2005) Gene set enrichment analysis: a knowledge-based approach for interpreting genome-wide expression profiles. Proceedings of the National Academy of Sciences of the United States of America 102:15545–15550
43. Singh D, Febbo PG, Ross K, et al (2002) Gene expression correlates of clinical prostate cancer behavior. Cancer Cell 1:203–209
44. Borovecki F, Lovrecic L, Zhou J, et al (2005) Genome-wide expression profiling of human blood reveals biomarkers for Huntington's disease. Proceedings of the National Academy of Sciences of the United States of America 102:11023–11028
45. Tian E, Zhan F, Walker R, et al (2003) The Role of the Wnt-Signaling Antagonist DKK1 in the Development of Osteolytic Lesions in Multiple Myeloma. New England Journal of Medicine 349:2483–2494
46. Li T, Zhang C, Ogihara M (2004) A comparative study of feature selection and multiclass classfication methods for tissue classification based on gene expression. Bioinformatics (Oxford, England) 20:2429–37. https://doi.org/10.1093/bioinformatics/bth267
47. Pedregosa F, Varoquaux G, Gramfort A, et al (2012) Scikit-learn: Machine Learning in Python
48. Calzolari M (2019) manuel-calzolari/sklearn-genetic: sklearn-genetic 0.2. Zenodo
49. Soufan O, Kleftogiannis D, Kalnis P, Bajic VB (2015) DWFS: A Wrapper Feature Selection Tool Based on a Parallel Genetic Algorithm. PLOS ONE 10:e0117988. https://doi.org/10.1371/journal.pone.0117988
50. Syafrudin M, Alfian G, Fitriyani NL, et al (2020) A Self-Care Prediction Model for Children with Disability Based on Genetic Algorithm and Extreme Gradient Boosting. Mathematics 8:1590. https://doi.org/10.3390/math8091590
51. Hall MA (1999) Correlation-based Feature Selection for Machine Learning. 198

52. Urbanowicz RJ, Olson RS, Schmitt P, et al (2017) Benchmarking relief-based feature selection methods
53. Wilcoxon F (1945) Individual Comparisons by Ranking Methods. Biometrics Bulletin 1:80–83. https://doi.org/10.2307/3001968
54. Pratt JW (1959) Remarks on Zeros and Ties in the Wilcoxon Signed Rank Procedures. Journal of the American Statistical Association 54:655–667. https://doi.org/10.1080/01621459.1959.10501526
55. Barot RK, Shitole SC, Bhagat N, et al (2016) Therapeutic effect of 0.1% Tacrolimus Eye Ointment in Allergic Ocular Diseases. J Clin Diagn Res 10:NC05–NC09. https://doi.org/10.7860/JCDR/2016/17847.7978
56. Maino P, Presilla S, Colli Franzone PA, et al (2018) Radiation Dose Exposure for Lumbar Transforaminal Epidural Steroid Injections and Facet Joint Blocks Under CT vs. Fluoroscopic Guidance. Pain Pract 18:798–804. https://doi.org/10.1111/papr.12677